\documentclass[10pt,twocolumn,letterpaper]{article}

\usepackage{cvpr}
\usepackage{times}
\usepackage{graphicx}
\usepackage{amsmath}
\usepackage{amssymb}
\usepackage{color}
\usepackage{graphics}
\usepackage{multirow}
\usepackage{subfig}
\usepackage{array}
\usepackage{colortbl}
\usepackage{cite}
\usepackage{diagbox}
\usepackage{rotating}
\usepackage{booktabs}
\usepackage{overpic}
\usepackage{contour}

\definecolor{mygray}{gray}{.92}

\newcommand{\figref}[1]{Fig. \ref{#1}}

\newcommand{\tabincell}[2]{\begin{tabular}{@{}#1@{}}#2\end{tabular}}

\usepackage[pagebackref=false,breaklinks=true,letterpaper=true,colorlinks,bookmarks=false]{hyperref}

\cvprfinalcopy 

\def\cvprPaperID{1303} 

\ifcvprfinal\fi

\def\ourmodel{\emph{JL-DCF}}
\graphicspath{{./Imgs/}}
\DeclareGraphicsExtensions{.jpg,.pdf,.png}
\begin{document}

\title{JL-DCF: Joint Learning and Densely-Cooperative Fusion \\ Framework for RGB-D Salient Object Detection}

\author{
Keren  Fu$^{1}$\quad
Deng-Ping Fan$^{2,3,}$\thanks{Corresponding author: Deng-Ping Fan \emph{(dengpfan@gmail.com)}}\quad
Ge-Peng Ji$^{4}$\quad
Qijun Zhao$^1$\\
$^1$ College of Computer Science, Sichuan University\quad
$^2$ College of CS, Nankai University \quad\\
$^3$ Inception Institute of Artificial Intelligence \quad  
$^4$ School of Computer Science, Wuhan University \\
{\tt \small \href{http://dpfan.net/JLDCF/}{http://dpfan.net/JLDCF/}}
}

\maketitle

\begin{abstract}
   This paper proposes a novel joint learning and densely-cooperative
   fusion (\textbf{\emph{\ourmodel}}) architecture for RGB-D salient object detection.
   Existing  models usually treat RGB and depth as independent
   information and design separate networks for feature extraction from each.
   Such schemes can easily be constrained by a limited amount of training
   data or over-reliance on an elaborately-designed training process.
   In contrast, our \emph{\ourmodel}~learns from both RGB and depth inputs through a Siamese network. To this end, we propose two effective components: joint learning
   (JL), and densely-cooperative fusion (DCF).
   The JL module provides robust saliency feature learning, while the latter is introduced for complementary feature discovery.
   Comprehensive experiments on four popular metrics show that the designed
   framework yields a robust RGB-D saliency detector with good generalization.
   As a result, JL-DCF significantly advances the top-1 D3Net model by an average of $\sim$1.9\%
   (S-measure) across six challenging datasets,
   showing that the proposed framework offers a potential solution
   for real-world applications and could provide more insight into the cross-modality complementarity task. The code will be available at
\href{https://github.com/kerenfu/JLDCF}{https://github.com/kerenfu/JLDCF/}.
\end{abstract}

\vspace{-5pt}
\section{Introduction}
\vspace{-5pt}

Salient object detection (SOD) aims at detecting the objects in a scene that humans would naturally focus on \cite{cheng2015global,borji2014salient,zhao2019egnet}. It has many useful applications, including
object segmentation and recognition \cite{Liu2012,ye2017salient,zhoucvpr2020,jerripothula2016image,
Rutishauser2004,han2006unsupervised}, image/video compression \cite{Guo2010},
video detection/summarization \cite{Ma2005,fan2019shifting}, content-based image editing
\cite{wang2017deep,Stentiford2007,Marchesotti2009,Ding2011,Goferman2010}, informative common object
discovery \cite{zhang2016detection,zhang2017co}, and image retrieval \cite{Chen2009,Gao2015,liu2013model}.
Many SOD models have been developed under the assumption that the inputs are individual
RGB/color images \cite{wang2016correspondence,zhang2017amulet,zhang2017learning,zhang2018progressive,feng2019attentive,piao2019deep} or sequences \cite{wang2019learning,wang2019zero,song2018pyramid,wang2019revisiting}. As depth cameras such as Kinect and RealSense become more and more
popular, SOD from RGB-D inputs (``D'' refers to depth) is emerging as an attractive
topic. Although a number of prior works have tried to explore the role of depth
in saliency analysis, several issues remain:

\begin{figure}
\includegraphics[width=0.46\textwidth]{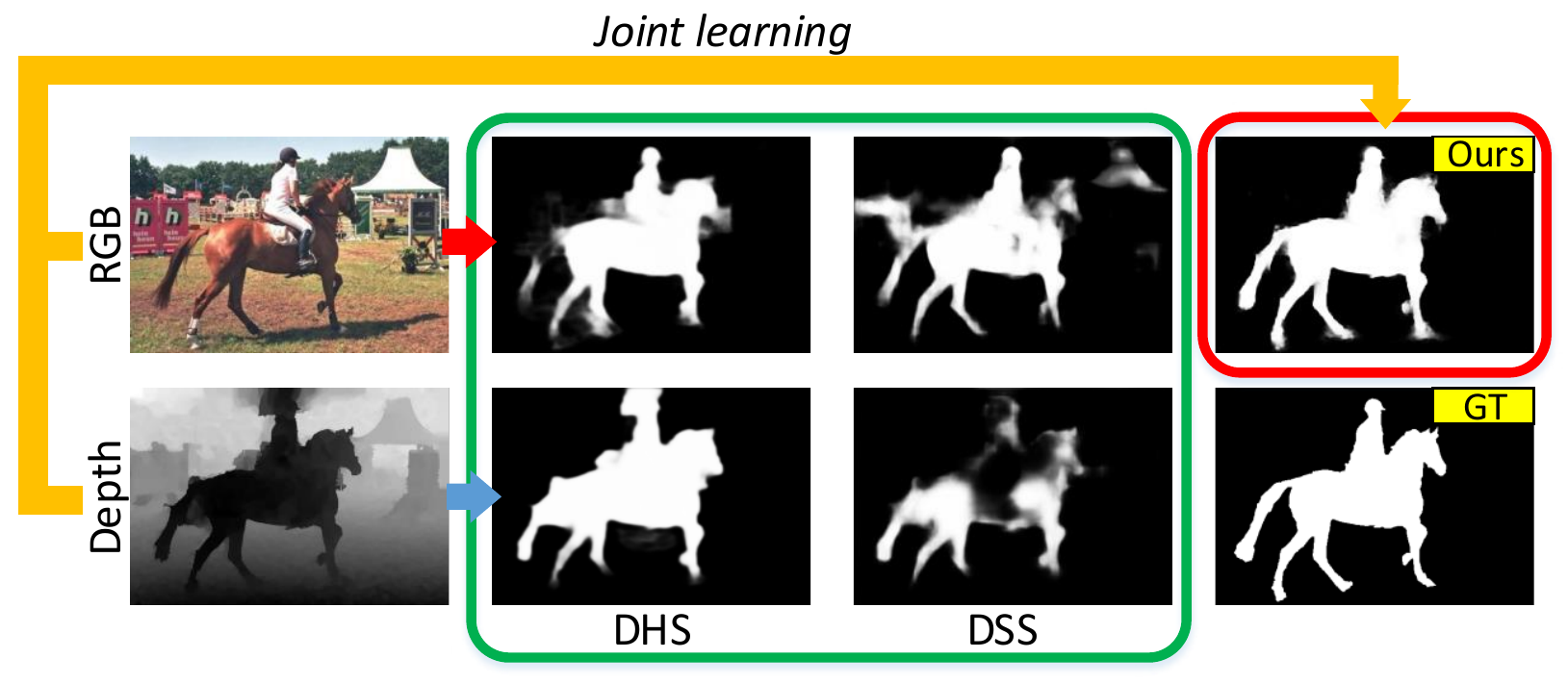} \vspace{-0.4cm}
\caption{\small Applying deep saliency models DHS \cite{liu2016dhsnet} and
DSS \cite{hou2019deeply}, which are fed with an RGB image (1$^{st}$ row) or a depth map (2$^{nd}$ row).
Both of the models are trained on a single RGB modality. By contrast, our \ourmodel~considers both modalities and thus generates better results (last column).}
\label{fig_motivation}
\vspace{-0.5cm}
\end{figure}

\textbf{(i) Deep-based RGB-D SOD methods are still under-explored:}
Despite more than one hundred papers on RGB SOD models being published since 2015 \cite{Fan2018SOC,wang2019salient,wang2019iterative,wang2018salient,wang2019salient2}, there are only a few deep
learning-based works focusing on RGB-D SOD. The first model utilizing
convolutional neural networks (CNNs) for RGB-D SOD \cite{qu2017rgbd}, which adopts a shallow CNN as the saliency map integration model, was described in 2017. Since then, only a dozen deep models have been proposed, as summarized in \cite{fan2019rethinking,Zhang2020UCNet}, leaving large room for further improvement in performance.

\textbf{(ii) Less effective feature extraction and fusion:}
Most learning-based models fuse features of different modalities
either by early-fusion \cite{song2017depth,liu2019salient,huang2018rgbd,
fan2019rethinking} or late-fusion \cite{han2017cnns,wang2019adaptive}.
Although these two simple strategies have achieved encouraging
progress in this field in the past (as pointed out in \cite{chen2018progressively}), they face challenges in either extracting representative multi-modal
features or effectively fusing them. While other works have adopted a middle-fusion strategy \cite{chen2018progressively,zhu2019pdnet,chen2019three},
which conducts independent feature extraction and fusion using individual CNNs, their
sophisticated network architectures and large number of parameters require an elaborately-designed training process and large amount of training data.
Unfortunately, high-quality depth maps are still sparse \cite{zhao2019contrast},
which may lead to sub-optimal solutions of deep learning-based models.

\noindent\textbf{Motivation.} To tackle RGB-D SOD, we propose a novel joint learning and densely-cooperative
fusion (\ourmodel) architecture that
outperforms all existing deep learning-based techniques.
Our method adopts the middle-fusion strategy mentioned above.
However, different from previous works which conduct independent feature
extraction from RGB and depth views, \ourmodel~effectively extracts deep
hierarchical features from RGB and depth inputs simultaneously, through a Siamese network (shared backbone).
The underlying motivation is that, although depth and RGB
images come from different modalities, they nevertheless share similar features/cues, such as strong figure-ground contrast \cite{niu2012leveraging,
peng2014rgbd,cheng2014depth}, closure of object contours \cite{feng2016local,
shigematsu2017learning}, and connectivity to image borders \cite{wang2017rgb,
liang2018stereoscopic}. This makes cross-modal transferring feasible, even for deep models. As evidenced in Fig. \ref{fig_motivation}, a model trained on a single RGB modality, like DHS \cite{liu2016dhsnet}, can sometimes perform well in the depth view. Nevertheless, a similar model, like DSS \cite{hou2019deeply}, could also fail in the depth view without proper adaption or transferring.

To the best of our knowledge, the proposed \ourmodel~scheme is \emph{the first to
leverage such transferability in deep models}, by treating a depth image as a
special case of a color image and employing a shared CNN for both RGB and depth
feature extraction. Additionally, we develop a densely-cooperative fusion strategy to
reasonably combine the learned features of different modalities.
This paper provides two main contributions:

\begin{itemize}
\vspace{-5pt}
\item We introduce a general framework for RGB-D SOD,
called \ourmodel, which consists of two sub-modules: joint learning and
densely-cooperative fusion. The key features of these two components are their robustness and effectiveness, which will be beneficial for future modeling in related multi-modality tasks in computer vision.
In particular, we advance the state-of-the-art (SOTA) by a significant average of $\sim$2\% (F-measure score) across six challenging datasets.

\vspace{-5pt}
\item We present a thorough evaluation of 14 SOTA methods~\cite{ju2014depth,feng2016local,cong2016saliency,song2017depth,
guo2016salient,qu2017rgbd,wang2019adaptive,han2017cnns,chen2019multi,
chen2018progressively,chen2019three,zhao2019contrast,fan2019rethinking,
Piao2019depth}, which is the largest-scale comparison in this field to date.
Besides, we conduct a comprehensive ablation study, including using different input sources, learning schemes, and feature fusion strategies, to demonstrate the effectiveness of \ourmodel.
Some interesting findings also encourage further research in this field.
\vspace{-0.2cm}
\end{itemize}

\vspace{-5pt}
\section{Related Work}\label{sec2}
\vspace{-5pt}
\noindent\textbf{Traditional.}
The pioneering work for RGB-D SOD was produced by Niu \emph{et al.} \cite{niu2012leveraging}, who introduced disparity contrast and domain knowledge into stereoscopic photography to measure stereo saliency. After Niu's work, various hand-crafted features/hypotheses originally applied for RGB SOD were extended to RGB-D, such as center-surround difference\cite{ju2014depth,guo2016salient}, contrast \cite{cheng2014depth,peng2014rgbd,cong2016saliency}, background enclosure \cite{feng2016local}, center/boundary prior \cite{cheng2014depth,liang2018stereoscopic,wang2017rgb,cong2019going}, compactness \cite{cong2016saliency,cong2019going}, or a combination of various saliency measures \cite{song2017depth}. All the above models rely heavily on heuristic hand-crafted features, resulting in limited generalizability in complex scenarios.

\begin{figure*}
\includegraphics[width=.98\textwidth]{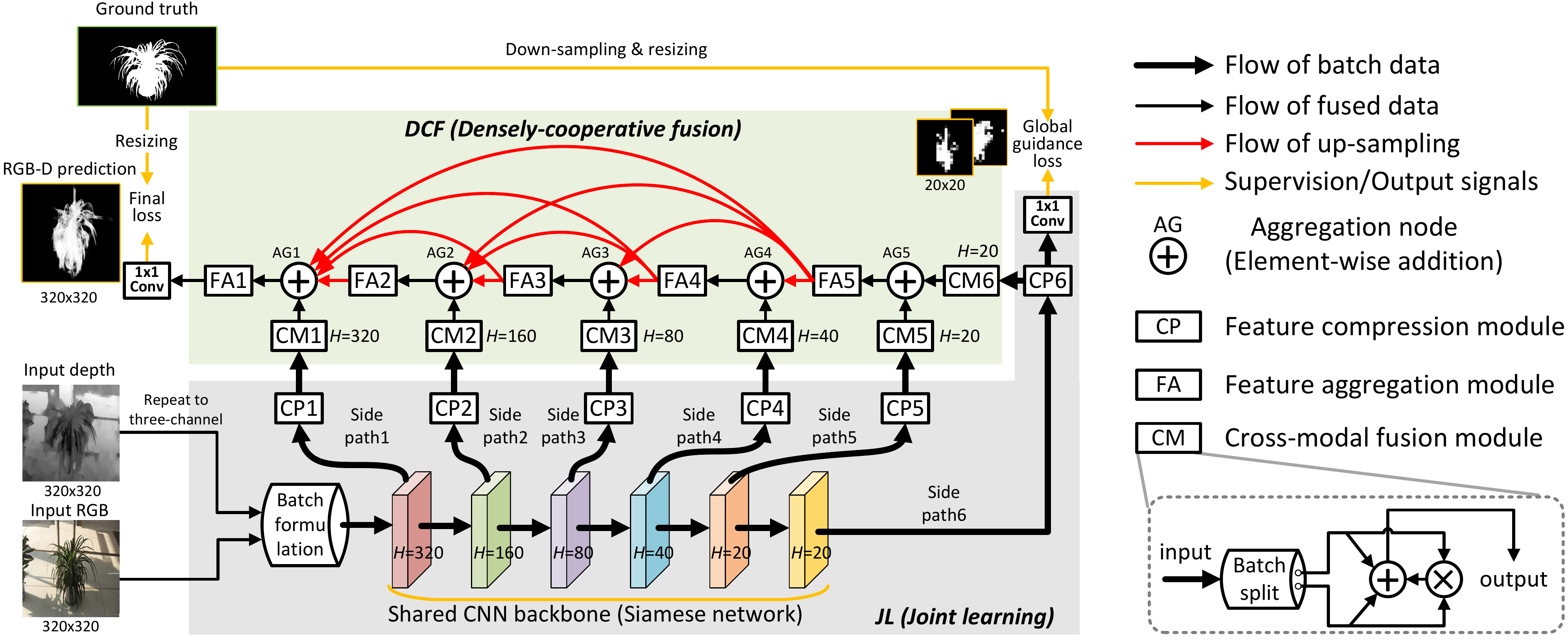} \vspace{-0.2cm}
\caption{\small Block diagram of the proposed \ourmodel~framework for RGB-D SOD. The
JL (joint learning) component is shown in gray, while the DCF (densely-cooperative fusion) component is shown in light green. CP1$\sim$CP6: Feature compression modules. FA1$\sim$FA6: Feature aggregation modules. CM1$\sim$CM6: Cross-modal fusion modules. ``\emph{H}'' denotes the spatial size of output feature maps on a particular stage. See Section \ref{sec3} for details.}
\label{fig_blockdiagram}
\vspace{-0.2cm}
\end{figure*}

\noindent\textbf{Deep-based.}
Recent advances in this field have been obtained by using deep learning and CNNs. Qu \emph{et al.} \cite{qu2017rgbd} first utilized a CNN to fuse different low-level saliency cues for judging the saliency confidence values of superpixels. Shigematsu \emph{et al.} \cite{shigematsu2017learning} extracted ten superpixel-based hand-crafted depth features capturing the background enclosure cue, depth contrast, and histogram distance. These features are fed to a CNN, whose output is shallowly fused with the RGB feature output to compute superpixel saliency.

A recent trend in this field is to exploit fully convolutional neural networks (FCNs) \cite{Long2017Fully}. Chen \emph{et al.} \cite{chen2018progressively} proposed a bottom-up/top-down architecture \cite{pinheiro2016learning}, which progressively performs cross-modal complementarity-aware fusion in its top-down pathway. Han \emph{et al.} \cite{han2017cnns} modified/extended the structure of the RGB-based deep neural network in order for it to be applicable for the depth view and then fused the deep representations of both views via a fully-connected layer. A three-stream attention-aware network was proposed in \cite{chen2019three}, which extracts hierarchical features from RGB and depth inputs through two separate streams. Features are then progressively combined and selected via attention-aware blocks in the third stream. A new multi-scale multi-path fusion network with cross-modal interactions was proposed in \cite{chen2019multi}. \cite{liu2019salient} and \cite{huang2018rgbd} formulated a four-channel input by concatenating RGB and depth. The input is later fed to a single-stream recurrent CNN and an FCN with short connections, respectively. \cite{zhu2019pdnet} employed a subsidiary network to obtain depth features and used them to enhance the intermediate representation in an encoder-decoder architecture. Zhao \emph{et al.} \cite{zhao2019contrast} proposed a model that generates a contrast-enhanced depth map, which is later used as a prior map for feature enhancement in subsequent fluid pyramid integration. Fan \emph{et al.} \cite{fan2019rethinking} constructed a new RGB-D dataset called the Salient Person (SIP) dataset, and introduced a depth-depurator network to judge whether a depth map should be concatenated with the RGB image to formulate an input signal.

Generally, as summarized by previous literature \cite{chen2018progressively,zhao2019contrast}, the above approaches can be divided into three categories: (a) Early-fusion \cite{song2017depth,liu2019salient,huang2018rgbd,fan2019rethinking}, (b) late-fusion \cite{han2017cnns,wang2019adaptive} and (c) middle-fusion  \cite{chen2018progressively,zhu2019pdnet,chen2019three,chen2019multi}.
Middle-fusion complements (a) and (b), since both feature-extraction and subsequent-fusion are handled by relatively deep CNNs. As a consequence, high-level concepts can be learnt from both modalities and complex integration rules can be mined. Besides, performing individual deep supervision for RGB and depth is straightforward. The proposed \ourmodel~scheme falls into the middle-fusion strategy.

However, unlike the aforementioned methods \cite{chen2018progressively,zhu2019pdnet,chen2019three,chen2019multi}, where the two feature extraction streams are independent, we propose to utilize a Siamese architecture \cite{chopra2005learning}, where both the network architecture and weights are shared. This results in two major benefits: 1) Cross-modal knowledge-sharing becomes straightforward via joint learning; 2) The model parameters are largely reduced as only one shared network is needed, leading to facilitated learning.

\vspace{-5pt}
\section{Methodology}\label{sec3}
\vspace{-5pt}

The overall architecture of the proposed \ourmodel~is shown in Fig. \ref{fig_blockdiagram}. It follows the classic bottom-up/top-down strategy \cite{pinheiro2016learning}. For illustrative purpose, Fig. \ref{fig_blockdiagram} depicts an example backbone with six hierarchies that are common in the widely-used VGG \cite{Simonyan14c} and ResNet \cite{He2015Deep}. The architecture consists of a JL component and a DCF component. The JL component conducts joint learning for the two modalities using a Siamese network. It aims to discover the commonality between these two views from a ``model-based’’ perspective, since their information can be merged into model parameters via back-propagation. As seen in Fig. \ref{fig_blockdiagram}, the hierarchical features jointly learned by the backbone are then fed to the subsequent DCF component. DCF is dedicated to feature fusion and its layers are constructed in a densely-cooperative way. In this sense, the complementarity between RGB and depth modalities can be explored from a ``feature-based'' perspective. To perform cross-view feature fusion, in the DCF component, we elaborately design a cross-modal fusion module (CM module in Fig. \ref{fig_blockdiagram}). Details about \ourmodel~will be given in the following sections.

\vspace{-5pt}
\subsection{Joint Learning (JL)}\label{sec32}
\vspace{-5pt}
As shown in Fig. \ref{fig_blockdiagram} (gray part), the inputs of the JL component are an RGB image together with its corresponding depth map. We first normalize the depth map into intervals [0, 255] and then convert it to a three-channel map through color mapping. In our implementation, we use the naive gray color mapping, which is equivalent to replicating the single channel map into three channels. Note that other color mapping \cite{al2016creating} or transformations, like the mean used in \cite{han2017cnns}, could also be considered for generating the three-channel representation. Next, the three-channel RGB image and transformed depth map are concatenated to formulate a \emph{batch}, so that the subsequent CNN backbone can perform parallel processing. Note that, unlike previous early-fusion schemes aforementioned, which often concatenate the RGB and depth inputs in the 3$^{rd}$ channel dimension, our scheme concatenates in the 4$^{th}$ dimension, often called the batch dimension. For example, in our case a transformed $320\times320\times3$ depth and a $320\times320\times3$ RGB map will formulate a batch of size $320\times320\times3\times2$, rather than $320\times320\times6$. 

The hierarchical features from the shared CNN backbone are then leveraged in a side-output way like \cite{hou2019deeply}. Since the side-output features have varied resolutions and channel numbers (usually the deeper, the more channels), we first employ a set of CP modules (CP1$\sim$CP6 in Fig. \ref{fig_blockdiagram}) to compress the side-output features to an identical, smaller number, denoted as $k$. We do this for the following two reasons: (1) Using a large number of feature channels for subsequent decoding is memory and computationally expensive and (2) Unifying the number of feature channels facilitates various element-wise operations. Note that, here, the outputs from our CP modules are still batches, which are denoted as the thicker black arrows in Fig. \ref{fig_blockdiagram}.

Coarse localization can provide the basis for the following top-down refinement \cite{pinheiro2016learning}.
In addition, jointly learning the coarse localization guides the shared CNN to learn to extract independent hierarchical features
from the RGB and depth views simultaneously.
In order to enable the CNN backbone to coarsely locate the targets from both the RGB and depth views, we apply deep supervision to the JL component in the last hierarchy.
To conduct this, as shown in Fig. \ref{fig_blockdiagram}, we add a $(1 \times 1,1)$ convolutional layer on the CP6 module to achieve coarse prediction. The depth and RGB-associated outputs are supervised by the down-sampled ground truth map. The generated loss in this stage is called the global guidance loss $\mathcal{L}_{g}$.


\vspace{-5pt}
\subsection{Densely-cooperative Fusion (DCF)}\label{sec33}
\vspace{-5pt}
As shown in Fig. \ref{fig_blockdiagram} (light green part), the output batch features from the CP modules contain depth and RGB information. They are fed to the DCF component, which can be deemed a decoder that performs multi-scale cross-modal fusion. Firstly, we design a CM (cross-modal fusion) module to split and then merge the batch features (Fig. \ref{fig_blockdiagram}, bottom-right). This module first splits the batch data and then conducts ``addition and multiplication'' feature fusion, which we call \emph{cooperative fusion}. Mathematically, let a batch feature be denoted by $\{{X}_{rgb}, {X}_d\}$, where ${X}_{rgb}$, ${X}_d$ represent the RGB and depth parts, each with $k$ channels, respectively. The CM module conducts the fusion as:

\vspace{-0.4cm}
\begin{equation} \label{equ_cm}
CM(\{{X}_{rgb}, {X}_d\})={X}_{rgb} \oplus {X}_d \oplus ( {X}_{rgb} \otimes {X}_d),
\end{equation}

\noindent where ``$\oplus$'' and ``$\otimes$'' denote element-wise addition and multiplication. The blended features output from the CM modules are still made up of $k$ channels. Compared to element-wise addition ``$\oplus$'', which exploits \emph{feature complementarity}, element-wise multiplication ``$\otimes$'' puts more emphasis on \emph{commonality}. These two properties are generally important in cross-view fusion.

One may argue that such a CM module could be replaced by channel concatenation, which generates $2k$-channel concatenated features. However, we find such a choice tends to result in the learning process being trapped in a local optimum, where it becomes biased towards only RGB information. The reason seems to be that the channel concatenation does indeed involve feature selection rather than explicit feature fusion. This leads to degraded learning outcomes, where only RGB features dominate the final prediction. Note that, as will be shown in Section \ref{sec44}, solely using RGB input can also achieve fairly good performance in the proposed framework. Comparisons between our CM modules and concatenation will be given in Section \ref{sec44}. 

\begin{figure}
\includegraphics[width=0.48\textwidth]{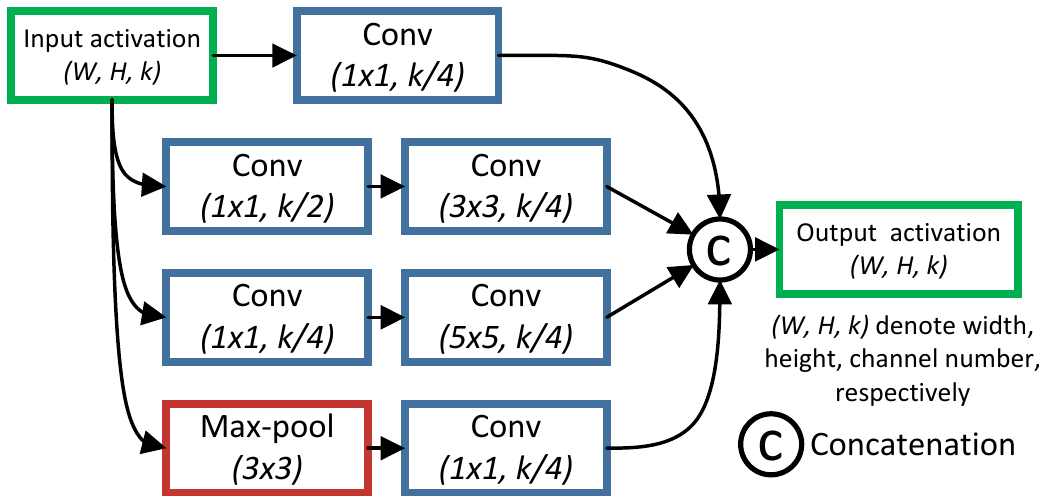} \vspace{-0.7cm}
\caption{\small Inception structure used for the FA modules in Fig. \ref{fig_blockdiagram}. All convolutional layers and max-pooling layers have stride 1, therefore maintaining spatial feature sizes. Unlike the original Inception module \cite{szegedy2015going}, we adapt it to have the same input/output channel number $k$.}
\label{fig_inception}
\vspace{-0.2cm}
\end{figure}

As shown in Fig. \ref{fig_blockdiagram}, the fused features from CM1$\sim$CM6 are fed to a decoder augmented with a dense connection \cite{huang2017densely}. Using the dense connection promotes the blending of depth and RGB features at various scales. Therefore, unlike the traditional UNet-like decoder \cite{ronneberger2015u}, an aggregation module FA takes inputs from all levels deeper than itself. Specifically, FA denotes a feature aggregation module performing non-linear aggregation. To this end, we use the Inception module \cite{szegedy2015going} shown in Fig. \ref{fig_inception}, which performs multi-level convolutions with filter size $1\times1,3\times3,5\times5$, and max-pooling. Note that the FA module in our framework is flexible. Other modules may also be considered in the future to improve the performance.

Finally, the FA module with the finest features is denoted as FA1, whose output is then fed to a $(1 \times 1,1)$ convolutional layer to generate the final activation and then  ultimately the saliency map. This final prediction is supervised by the resized ground truth (GT) map during training. We denote the loss generated in this stage as $\mathcal{L}_{f}$.

\begin{table*}[t!]
    \renewcommand{\arraystretch}{1.0}
    \caption{\small Quantitative measures: S-measure ($S_\alpha$) \cite{Fan2017}, max F-measure ($F_{\beta}^{\textrm{max}}$) \cite{Borji2015TIP}, max E-measure ($E_{\phi}^{\textrm{max}}$) \cite{fan2018enhanced} and MAE ($M$) \cite{Perazzi2012} of SOTA methods and the proposed \ourmodel~on six RGB-D datasets. The best performance is highlighted in \textbf{bold}.}\label{tab_sota}
    \vspace{-0.3cm}
   \centering
    \footnotesize
    \setlength{\tabcolsep}{1.3mm}
    \begin{tabular}{p{0.8mm}p{0.8mm}r||c|c|c|c|c||c|c|c|c|c|c|c|c|c|c}
    \hline
            && Metric  & \tabincell{c}{ACSD\\\cite{ju2014depth}} & \tabincell{c}{LBE\\\cite{feng2016local}} & \tabincell{c}{DCMC\\\cite{cong2016saliency}}  & \tabincell{c}{MDSF\\\cite{song2017depth}} & \tabincell{c}{SE\\\cite{guo2016salient}} & \tabincell{c}{DF\\\cite{qu2017rgbd}} & \tabincell{c}{AFNet\\\cite{wang2019adaptive}} & \tabincell{c}{CTMF\\\cite{han2017cnns}} & \tabincell{c}{MMCI\\\cite{chen2019multi}} & \tabincell{c}{PCF\\\cite{chen2018progressively}} & \tabincell{c}{TANet\\\cite{chen2019three}} & \tabincell{c}{CPFP\\\cite{zhao2019contrast}} & \tabincell{c}{DMRA\\\cite{Piao2019depth}}& \tabincell{c}{D3Net\\\cite{fan2019rethinking}}& \tabincell{c}{\ourmodel\\Ours}\\
    \hline
    \hline
          \multirow{4}{*}{\begin{sideways}\textit{NJU2K}\end{sideways}} & \multirow{4}{*}{\begin{sideways}\cite{ju2014depth}\end{sideways}} & $S_\alpha\uparrow$       &0.699&0.695&0.686&0.748&0.664&0.763&0.772&0.849&0.858&0.877&0.878&0.879&0.886&0.895&\textbf{0.903}\\
                                                                        && $F_{\beta}^{\textrm{max}}\uparrow$       &0.711&0.748&0.715&0.775&0.748&0.804&0.775&0.845&0.852&0.872&0.874&0.877&0.886&0.889&\textbf{0.903}\\
                                                                        && $E_{\phi}^{\textrm{max}}\uparrow$       &0.803&0.803&0.799&0.838&0.813&0.864&0.853&0.913&0.915&0.924&0.925&0.926&0.927&0.932&\textbf{0.944}\\
                                                                        && $M\downarrow$       &0.202&0.153&0.172&0.157&0.169&0.141&0.100&0.085&0.079&0.059&0.060&0.053&0.051&0.051&\textbf{0.043}\\
    \hline
        \multirow{4}{*}{\begin{sideways}\textit{NLPR}\end{sideways}} & \multirow{4}{*}{\begin{sideways}\cite{peng2014rgbd}\end{sideways}}& $S_\alpha\uparrow$
        &0.673&0.762&0.724&0.805&0.756&0.802&0.799&0.860&0.856&0.874&0.886&0.888&0.899&0.906&\textbf{0.925}\\
                                                                        && $F_{\beta}^{\textrm{max}}\uparrow$      &0.607&0.745&0.648&0.793&0.713&0.778&0.771&0.825&0.815&0.841&0.863&0.867&0.879&0.885&\textbf{0.916}\\
                                                                        && $E_{\phi}^{\textrm{max}}\uparrow$     &0.780&0.855&0.793&0.885&0.847&0.880&0.879&0.929&0.913&0.925&0.941&0.932&0.947&0.946&\textbf{0.962}\\
                                                                        && $M\downarrow$       &0.179&0.081&0.117&0.095&0.091&0.085&0.058&0.056&0.059&0.044&0.041&0.036&0.031&0.034&\textbf{0.022}\\

    \hline
        \multirow{4}{*}{\begin{sideways}\textit{STERE}\end{sideways}}& \multirow{4}{*}{\begin{sideways}\cite{niu2012leveraging}\end{sideways}} & $S_\alpha\uparrow$ &0.692&0.660&0.731&0.728&0.708&0.757&0.825&0.848&0.873&0.875&0.871&0.879&0.886&0.891&\textbf{0.905}\\
                                                                        && $F_{\beta}^{\textrm{max}}\uparrow$       &0.669&0.633&0.740&0.719&0.755&0.757&0.823&0.831&0.863&0.860&0.861&0.874&0.886&0.881&\textbf{0.901}\\
                                                                        && $E_{\phi}^{\textrm{max}}\uparrow$     &0.806&0.787&0.819&0.809&0.846&0.847&0.887&0.912&0.927&0.925&0.923&0.925&0.938&0.930&\textbf{0.946}\\
                                                                        && $M\downarrow$       &0.200&0.250&0.148&0.176&0.143&0.141&0.075&0.086&0.068&0.064&0.060&0.051&0.047&0.054&\textbf{0.042}\\
    \hline
        \multirow{4}{*}{\begin{sideways}\textit{RGBD135}\end{sideways}} & \multirow{4}{*}{\begin{sideways}\cite{cheng2014depth}\end{sideways}}  & $S_\alpha\uparrow$       &0.728&0.703&0.707&0.741&0.741&0.752&0.770&0.863&0.848&0.842&0.858&0.872&0.900&0.904&\textbf{0.929}\\
                                                                        && $F_{\beta}^{\textrm{max}}\uparrow$     &0.756&0.788&0.666&0.746&0.741&0.766&0.728&0.844&0.822&0.804&0.827&0.846&0.888&0.885&\textbf{0.919}\\
                                                                        && $E_{\phi}^{\textrm{max}}\uparrow$       &0.850&0.890&0.773&0.851&0.856&0.870&0.881&0.932&0.928&0.893&0.910&0.923&0.943&0.946&\textbf{0.968}\\
                                                                        && $M\downarrow$       &0.169&0.208&0.111&0.122&0.090&0.093&0.068&0.055&0.065&0.049&0.046&0.038&0.030&0.030&\textbf{0.022}\\
                                                                                                                                                                                                      \hline
        \multirow{4}{*}{\begin{sideways}\textit{LFSD}\end{sideways}} & \multirow{4}{*}{\begin{sideways}\cite{li2014saliency}\end{sideways}}  & $S_\alpha\uparrow$       &0.727&0.729&0.746&0.694&0.692&0.783&0.730&0.788&0.779&0.786&0.794&0.820&0.839&0.824&\textbf{0.854}\\
                                                                        && $F_{\beta}^{\textrm{max}}\uparrow$      &0.763&0.722&0.813&0.779&0.786&0.813&0.740&0.787&0.767&0.775&0.792&0.821&0.852&0.815&\textbf{0.862}\\
                                                                        && $E_{\phi}^{\textrm{max}}\uparrow$       &0.829&0.797&0.849&0.819&0.832&0.857&0.807&0.857&0.831&0.827&0.840&0.864&0.893&0.856&\textbf{0.893}\\
                                                                        && $M\downarrow$       &0.195&0.214&0.162&0.197&0.174&0.146&0.141&0.127&0.139&0.119&0.118&0.095&0.083&0.106&\textbf{0.078}\\
     \hline
        \multirow{4}{*}{\begin{sideways}\textit{SIP}\end{sideways}} & \multirow{4}{*}{\begin{sideways}\cite{fan2019rethinking}\end{sideways}} & $S_\alpha\uparrow$       &0.732&0.727&0.683&0.717&0.628&0.653&0.720&0.716&0.833&0.842&0.835&0.850&0.806&0.864&\textbf{0.879}\\
                                                                        && $F_{\beta}^{\textrm{max}}\uparrow$       &0.763&0.751&0.618&0.698&0.661&0.657&0.712&0.694&0.818&0.838&0.830&0.851&0.821&0.862&\textbf{0.885}\\
                                                                        && $E_{\phi}^{\textrm{max}}\uparrow$       &0.838&0.853&0.743&0.798&0.771&0.759&0.819&0.829&0.897&0.901&0.895&0.903&0.875&0.910&\textbf{0.923}\\                                                                     && $M\downarrow$       &0.172&0.200&0.186&0.167&0.164&0.185&0.118&0.139&0.086&0.071&0.075&0.064&0.085&0.063&\textbf{0.051}\\
\hline
\end{tabular}
\vspace{-0.4cm}
\end{table*}

\vspace{-5pt}
\subsection{Loss Function}\label{sec34}
\vspace{-5pt}
The overall loss function of our scheme is composed of the global guidance loss $\mathcal{L}_{g}$ and final loss $\mathcal{L}_{f}$. Assume that $G$ denotes supervision from the ground truth, $S^c_{rgb}$ and $S^{c}_{d}$ denote the coarse prediction maps contained in the batch after module CP6, and $S^{f}$ is the final prediction after module FA1. The overall loss function is defined as:

\vspace{-0.4cm}
\begin{equation} \label{equ_loss}
\mathcal{L}_{total}=\mathcal{L}_{f}(S^{f}, G) + \lambda \sum_{x \in \{ rgb, d\}}\mathcal{L}_{g}(S^c_{x}, G),
\end{equation}

\noindent where $\lambda$ balances the emphasis of global guidance, and we adopt the widely used cross-entropy loss for $\mathcal{L}_{g}$ and $\mathcal{L}_{f}$ as:

\vspace{-0.4cm}
\begin{equation} \label{equ_cel}
\mathcal{L}(S,G)=-\sum_{i}[G_i \log (S_i) + (1-G_i) \log (1-S_i)],
\end{equation}

\noindent where $i$ denotes pixel index, and $S \in \{S^c_{rgb}, S^c_{d}, S^{f}\}$.


\vspace{-5pt}
\section{Experiments}\label{sec4}
\vspace{-5pt}
\subsection{Datasets and Metrics}\label{sec41}
\vspace{-5pt}
Experiments are conducted on six public RGB-D benchmark datasets: NJU2K \cite{ju2014depth} (2000 samples), NLPR \cite{peng2014rgbd} (1000 samples), STERE \cite{niu2012leveraging} (1000 samples), RGBD135 \cite{cheng2014depth} (135 samples), LFSD \cite{li2014saliency} (100 samples), and SIP \cite{fan2019rethinking} (929 samples). Following \cite{zhao2019contrast}, we choose the same 700 samples from NLPR and 1500 samples from NJU2K to train our algorithms. The remaining samples are used for testing. For fair comparisons, we apply the model trained on this training set to other datasets. For evaluation, we adopt four widely used metrics, namely 
S-measure ($S_\alpha$) \cite{Fan2017,zhao2019contrast}, maximum F-measure ($F_{\beta}^{\textrm{max}}$) \cite{Borji2015TIP,hou2019deeply}, maximum E-measure ($E_{\phi}^{\textrm{max}}$) \cite{fan2018enhanced,fan2019rethinking}, and MAE ($M$) \cite{Perazzi2012,Borji2015TIP}. The definitions for these metrics are omitted here and readers are referred to the related papers. Note that, since the E-measure metric was originally proposed in \cite{fan2018enhanced} for evaluating binary maps, to extend it for comparing a non-binary saliency map against a binary ground truth map, we follow a similar strategy to $F_{\beta}^{\textrm{max}}$. Specifically, we first binarize a saliency map into a series of foreground maps using all possible threshold values in [0, 255], and then report the maximum E-measure among them.

\vspace{-5pt}
\subsection{Implementation Details}\label{sec42}
\vspace{-5pt}

The proposed \ourmodel~scheme is generally independent from the network backbone. In this work, we implement two versions of \ourmodel~based on VGG-16 \cite{Simonyan14c} and ResNet-101 \cite{He2015Deep}, respectively. We fix the input size of the network as $320 \times 320 \times 3$. Simple gray color mapping is adopted to convert a depth map into a three-channel map.

\textbf{VGG-16 configuration:} For the VGG-16 with the fully-connected layers removed and having 13 convolutional layers, the \emph{side} \emph{path1}$\sim$\emph{path6} are successively connected to \emph{conv1\_2}, \emph{conv2\_2}, \emph{conv3\_3}, \emph{conv4\_3}, \emph{conv5\_3}, and \emph{pool5}. Inspired by \cite{hou2019deeply}, we add two extra convolutional layers into \emph{side} \emph{path1}$\sim$\emph{path6}.
To augment the resolution of the coarsest feature maps from \emph{side} \emph{path6}, while at the same time preserving the receptive field, we let \emph{pool5} have a stride of 1 and instead use dilated convolution \cite{chen2017deeplab} with a rate of 2 for the two extra side convolutional layers.
In general, the coarsest features produced by our final modified VGG-16 backbone have a spatial size of $20 \times 20$, as shown in \figref{fig_blockdiagram}.

\textbf{ResNet-101 configuration:} Similar to the VGG-16 case above, the spatial size of the coarsest features produced by our modified ResNet-101 backbone is also $20 \times 20$. As the first convolutional layer of ResNet already has a stride of 2, the features from the shallowest level have a spatial size of $160 \times 160$. To obtain the full size ($320\times320$) features without trivial up-sampling, we borrow the \emph{conv1\_1} and \emph{conv1\_2} layers from VGG-16 for feature extraction. \emph{Side} \emph{path1}$\sim$\emph{path6} are connected to \emph{conv1\_2}, and \emph{conv1}, \emph{res2c}, \emph{res3b3}, \emph{res4b22}, \emph{res5c} of the ResNet-101, respectively. We also change the stride of the \emph{res5a} block from 2 to 1, but subsequently use dilated convolution with rate 2.

\begin{figure*}
\includegraphics[width=1\textwidth]{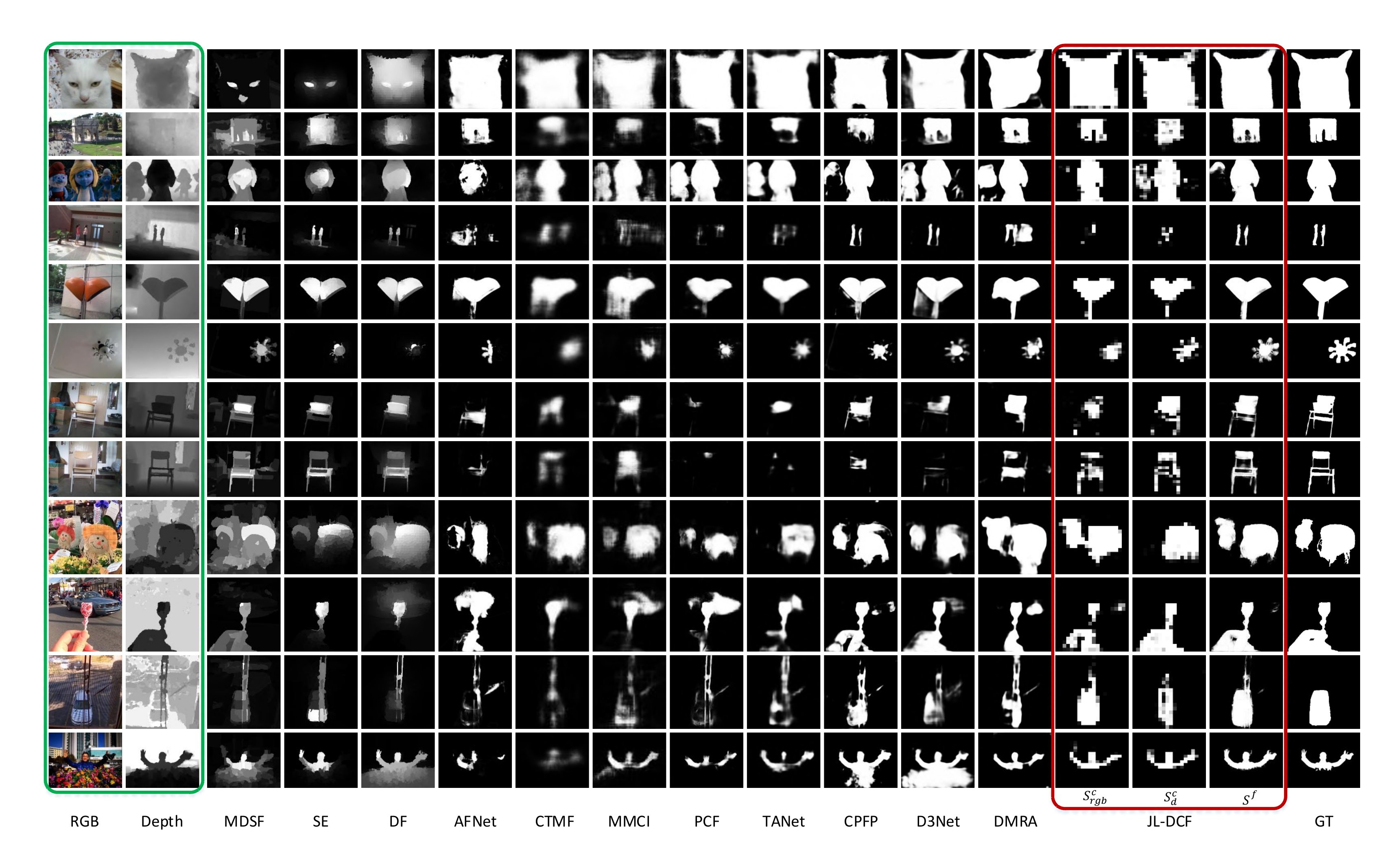} \vspace{-0.7cm}
\caption{\small Visual comparisons of \ourmodel~with SOTA RGB-D saliency models. The jointly learned coarse prediction maps ($S^c_{rgb}$ and $S^c_{d}$) from RGB and depth are also shown together with the final maps ($S^{f}$) of \ourmodel.}
\label{fig_sotavisual}
\vspace{-0.2cm}
\end{figure*}

\textbf{Decoder configuration:} All CP modules in Fig. \ref{fig_blockdiagram} are $3 \times 3$ convolutions with $k=64$ filters, and all FA modules are Inception modules. Up-sampling is achieved by simple bilinear interpolation. As depicted in Fig. \ref{fig_blockdiagram}, to align the feature sizes in the decoder, the output from an FA module is up-sampled by various factors. In an extreme case, the output from FA5 is up-sampled by a factor of $2$, $4$, $8$, and $16$. The final output from FA1 has a spatial size of $320 \times 320$, which is identical to the initial input.

\textbf{Training setup:} We implement \ourmodel~on Caffe \cite{jia2014caffe}. During training, the backbone~\cite{Simonyan14c,He2015Deep} is initialized by the pre-trained parameters of DSS \cite{hou2019deeply}, and other layers are randomly initialized. We fine-tune the entire network through end-to-end joint learning. Training data is augmented by mirror reflection to generate double the amount of data. The momentum parameter is set as 0.99, the learning rate is set to $lr=10^{-9}$, and the weight decay is 0.0005. The weight $\lambda$ in Eq. (\ref{equ_loss}) is set as 256 (=16$^2$) to balance the loss between the low- and high-resolution predictions. Stochastic Gradient Descent learning is adopted and accelerated by an NVIDIA 1080Ti GPU.
The training time is about 20 hours/18 hours for 40 epochs under the ResNet-101/VGG-16 configuration.

\vspace{-5pt}
\subsection{Comparisons to SOTAs}\label{sec43}
\vspace{-5pt}
We compare \ourmodel~(ResNet configuration) with 14 SOTA methods. Among the competitors, DF \cite{qu2017rgbd}, AFNet \cite{wang2019adaptive}, CTMF \cite{han2017cnns}, MMCI \cite{chen2019multi}, PCF \cite{chen2018progressively}, TANet \cite{chen2019three}, CPFP \cite{zhao2019contrast}, D3Net \cite{fan2019rethinking}, DMRA \cite{Piao2019depth} are recent deep learning-based methods, while ACSD \cite{ju2014depth}, LBE \cite{feng2016local}, DCMC \cite{cong2016saliency}, MDSF \cite{song2017depth}, SE \cite{guo2016salient} are traditional techniques using various hand-crafted features/hypotheses. Quantitative results are shown in Table \ref{tab_sota}. Notable performance gains of \ourmodel~over existing and recently proposed techniques, like CPFP\cite{zhao2019contrast}, D3Net\cite{fan2019rethinking} and DMRA\cite{Piao2019depth}, can be seen in all four metrics. This validates the consistent effectiveness of \ourmodel~and its generalizability.
Some visual examples are shown in Fig. \ref{fig_sotavisual}. \ourmodel~appears to be more effective at utilizing depth information for cross-modal compensation, making it better for detecting target objects in the RGB-D mode. Additionally, the deeply-supervised coarse predictions are listed in Fig. \ref{fig_sotavisual}. One can see that they provide basic object localization support for the subsequent cross-modal refinement, and our densely-cooperative fusion architecture learns an adaptive and ``image-dependent'' way of fusing such support with the hierarchical multi-view features. This proves that the fusion process does not degrade in either of the two views (RGB or depth), leading to boosted performance after fusion.

\begin{figure*}
\includegraphics[width=1.0\textwidth]{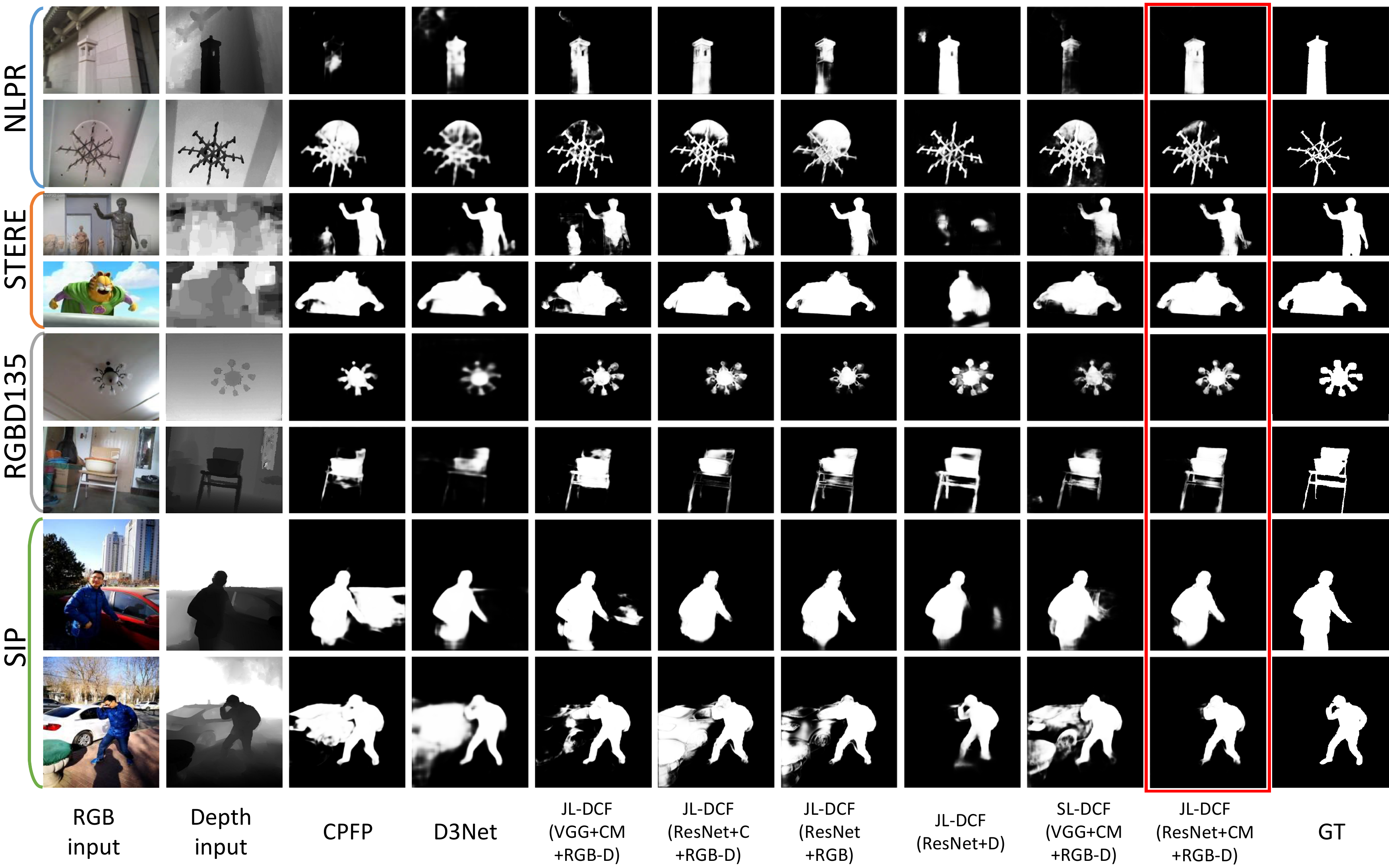} \vspace{-0.7cm}
\caption{\small Visual examples from NLPR, STERE, RGB135, SIP datasets for ablation studies. Generally, the full implementation of \ourmodel~(ResNet+CM+RGB-D, highlighted in the red box) achieves the closest results to the ground truth.}
\label{fig_ablationvisual}
\vspace{-0.2cm}
\end{figure*}

\vspace{-5pt}
\subsection{Ablation Studies}\label{sec44}
\vspace{-5pt}
We conduct thorough ablation studies by removing or replacing components from the full implementation of \ourmodel. We set the ResNet version of \ourmodel~as reference, and then compare various ablation experiments to it. We denote this reference version as ``\ourmodel~(ResNet+CM+RGB-D)'', where ``CM'' refers to the usage of CM modules and ``RGB-D'' refers to both RGB and depth inputs.

Firstly, to compare different backbones, a version ``\ourmodel~(VGG+CM+RGB-D)'' is trained by replacing the ResNet backbone with VGG, while keeping other settings unchanged. To validate the effectiveness of the adopted cooperative fusion modules, we train another version ``\ourmodel~(ResNet+C+RGB-D)'', by replacing the CM modules with a concatenation operation. To demonstrate the effectiveness of combining RGB and depth, we train two versions ``\ourmodel~(ResNet+RGB)'' and ``\ourmodel~(ResNet+D)'' respectively, where all the batch-related operations (such as CM modules) in Fig. \ref{fig_blockdiagram} are replaced with identity mappings, while all the other settings, including the dense decoder and deep supervision, are kept unchanged. Note that this validation is important to show that our network has learned complementary information by fusing RGB and depth. Lastly, to illustrate the benefit of joint learning, we train a scheme ``SL-DCF (VGG+CM+RGB-D)'' using two separate backbones for RGB and depth. ``SL'' stands for ``Separate Learning'', in contrast to the proposed ``Joint Learning''. In this test, we adopt VGG-16, which is smaller, since using two separate backbones leads to almost twice the overall model size.

Quantitative comparisons for various metrics are shown in Table \ref{tab_ablation}. Two SOTA methods CPFP \cite{zhao2019contrast} and D3Net \cite{fan2019rethinking} are listed for reference. Fig. \ref{fig_ablationvisual} shows visual ablation comparisons. Five different observations can be made:

\textbf{ResNet-101 $vs.$ VGG-16:} From the comparison between columns ``\texttt{A}'' and ``\texttt{B}'' in Table \ref{tab_ablation}, the superiority of the ResNet backbone over VGG-16 is evident, which is consistent with previous works. Note that the VGG version of our scheme still outperforms the leading methods CPFP (VGG-16 backbone) and D3Net (ResNet backbone).

\textbf{Effectiveness of CM modules:} Comparing columns ``\texttt{A}'' and ``\texttt{C}''  demonstrates that changing the CM modules into concatenation operations leads to a certain amount of degeneration. The underlying reason is that the whole network tends to bias its learning towards only RGB information, while ignoring depth, since it is able to achieve fairly good results (column ``\texttt{D}'') by doing so on the most datasets. Although concatenation is a popular way to fuse features, the learning may become easily trapped without appropriate guidance. In contrast, our CM modules perform the ``explicit fusion operation'' across RGB and depth modalities.

\begin{figure}
\centering
\includegraphics[width=0.38\textwidth]{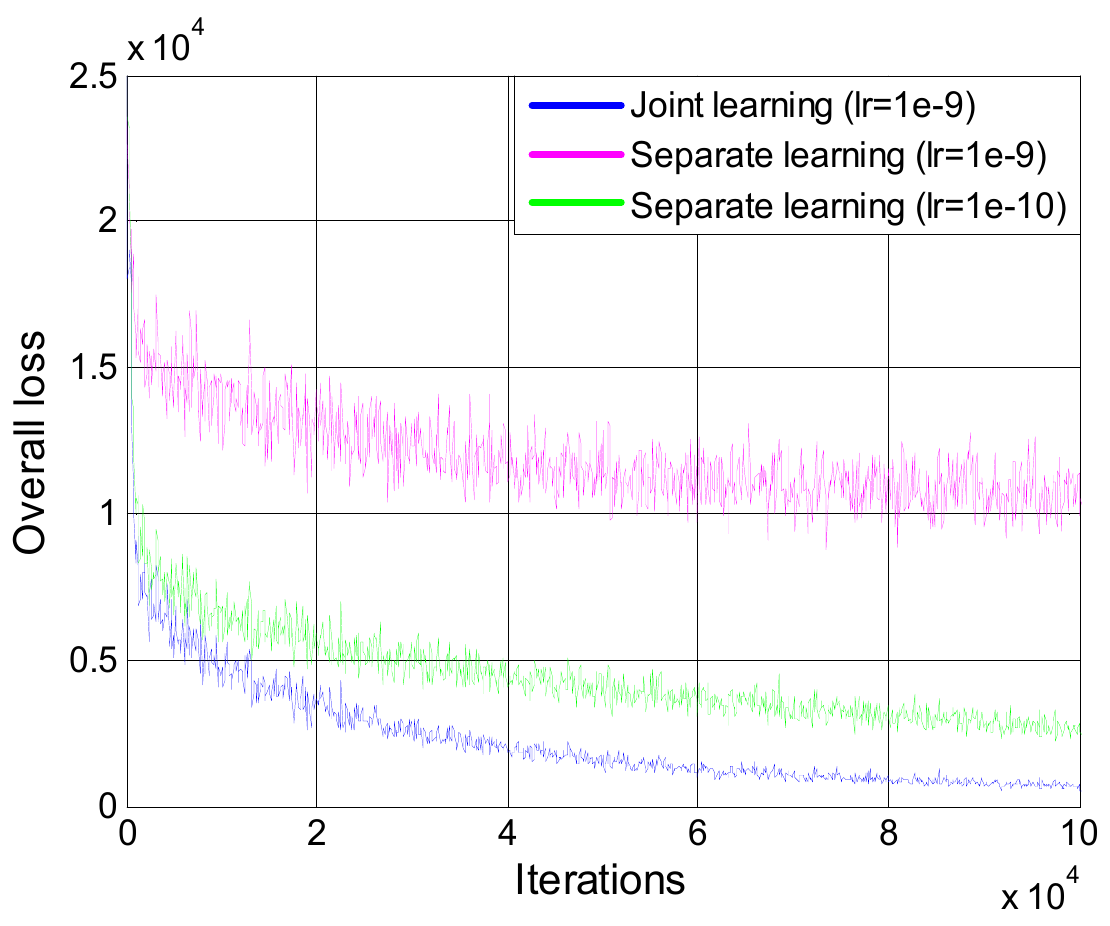}
\vspace{-10pt}
\caption{\small Learning curve comparison between joint learning (\ourmodel) and separate learning (SL-DCF).}
\label{fig_learningcurve}
\vspace{-0.2cm}
\end{figure}

\textbf{Combining RGB and depth:} The effectiveness of combining RGB and depth for boosting the performance is clearly validated by the consistent improvement over most datasets (compare column ``\texttt{A}''  with columns ``\texttt{D}'' and ``\texttt{E}''). The only exception is on STERE \cite{niu2012leveraging}, with the reason being that the quality of depth maps in this dataset is much worse compared to other datasets. Visual examples are shown in Fig. \ref{fig_ablationvisual}, in the 3$^{rd}$ and 4$^{th}$ rows. We find that many depth maps from STERE are too coarse and have very inaccurate object boundaries, misaligning with the true objects. Absorbing such unreliable depth information may, in turn, degrade the performance. Quantitative evidence can be seen in Table \ref{tab_ablation}, column ``\texttt{E}'' (STERE dataset), where solely using depth cues achieves much worse performance (about 16\%/20\% lower on $S_{\alpha}$/$F_{\beta}^{\textrm{max}}$ comparing to RGB) than on other datasets.

\begin{table}[t!]
    \renewcommand{\arraystretch}{0.95}
    \caption{\small Quantitative evaluation for ablation studies described in Section \ref{sec44}. For different configurations, ``\texttt{A}'': JL-DCF (ResNet+CM+RGB-D), ``\texttt{B}'': JL-DCF (VGG+CM+RGB-D), ``\texttt{C}'': JL-DCF (ResNet+C+RGB-D), ``\texttt{D}'': JL-DCF (ResNet+RGB), ``\texttt{E}'': JL-DCF (ResNet+D), ``\texttt{F}'': SL-DCF (VGG+CM+RGB-D).}\label{tab_ablation}
    \vspace{-0.2cm}
    \centering
    \footnotesize
    \setlength{\tabcolsep}{1.2mm}
    \begin{tabular}{p{0.8mm}p{0.8mm}r||c|c|c|c|c|c|c|c}
    \hline
          && Metric  & \tabincell{c}{CPFP} & \tabincell{c}{D3Net} & \tabincell{c}{\texttt{A}}  & \tabincell{c}{\texttt{B}} & \tabincell{c}{\texttt{C}} & \tabincell{c}{\texttt{D}} & \tabincell{c}{\texttt{E}} & \tabincell{c}{\texttt{F}}\\
    \hline
    \hline
          \multirow{4}{*}{\begin{sideways}\textit{NJU2K}\end{sideways}} & \multirow{4}{*}{\begin{sideways}\cite{ju2014depth}\end{sideways}}
          & $S_\alpha\uparrow$ &.878&.895&\textbf{.903}&.897
&.900&.895&.865&.886\\
                                                                        && $F_{\beta}^{\textrm{max}}\uparrow$       &.877&.889
 & \textbf{.903} & .899& .898 & .892& .863 & .883\\
                                                                        && $E_{\phi}^{\textrm{max}}\uparrow$       & .926 & .932& \textbf{.944}& .939& .937 & .937& .916 & .929\\
                                                                        && $M\downarrow$       & .053 & .051& \textbf{.043}& .044& .045 & .046& .063 & .053\\
    \hline
        \multirow{4}{*}{\begin{sideways}\textit{NLPR}\end{sideways}} & \multirow{4}{*}{\begin{sideways}\cite{peng2014rgbd}\end{sideways}}& $S_\alpha\uparrow$       & .888 & .906& \textbf{.925}& .920& .924 & .922& .873 & .901\\
                                                                        && $F_{\beta}^{\textrm{max}}\uparrow$       & .868& .885& \textbf{.916}& .907& .914 & .909& .843 & .881\\
                                                                        && $E_{\phi}^{\textrm{max}}\uparrow$       & .932 & .946& \textbf{.962}& .959& .961 & .957& .930 & .946\\
                                                                        && $M\downarrow$       & .036 & .034& \textbf{.022}& .026& .023 & .025& .041 & .033\\
    \hline
        \multirow{4}{*}{\begin{sideways}\textit{STERE}\end{sideways}}& \multirow{4}{*}{\begin{sideways}\cite{niu2012leveraging}\end{sideways}} & $S_\alpha\uparrow$       & .879& .891& .905& .894& .906 & \textbf{.909}& .744 & .886\\
                                                                        && $F_{\beta}^{\textrm{max}}\uparrow$       & .874& .881& \textbf{.901}& .889& .899& .901 & .708& .876\\
                                                                        && $E_{\phi}^{\textrm{max}}\uparrow$       & .925 & .930& \textbf{.946}& .938& .945 & .946& .834 & .931\\
                                                                        && $M\downarrow$       & .051 & .054& .042& .046& .041 & \textbf{.038}& .110 & .053\\
    \hline
        \multirow{4}{*}{\begin{sideways}\textit{RGBD135}\end{sideways}}& \multirow{4}{*}{\begin{sideways}\cite{cheng2014depth}\end{sideways}}   & $S_\alpha\uparrow$       & .872& .904& \textbf{.929}& .913& .916 & .903& .918 & .893\\
                                                                        && $F_{\beta}^{\textrm{max}}\uparrow$       & .846& .885& \textbf{.919}& .905& .906 & .894& .906 & .876\\
                                                                        && $E_{\phi}^{\textrm{max}}\uparrow$    & .923 & .946& \textbf{.968}& .955& .957 & .947& .967 & .950\\
                                                                        && $M\downarrow$       & .038 & .030& \textbf{.022}& .026& .025 & .027& .027 & .033\\
                                                                                                                                                                                                      \hline
        \multirow{4}{*}{\begin{sideways}\textit{LFSD}\end{sideways}}& \multirow{4}{*}{\begin{sideways}\cite{li2014saliency}\end{sideways}}   & $S_\alpha\uparrow$       & .820& .832& \textbf{.854}& .833& .852 & .845& .752 & .826\\
                                                                        && $F_{\beta}^{\textrm{max}}\uparrow$       & .821& .819& \textbf{.862}& .840& .854 & .846& .764 & .828\\
                                                                        && $E_{\phi}^{\textrm{max}}\uparrow$       & .864 & .864 & \textbf{.893}& .877& .893 & .889& .816 & .864\\
                                                                        && $M\downarrow$       & .095 & .099& \textbf{.078}& .091& .078 & .083& .126 & .101\\
     \hline
        \multirow{4}{*}{\begin{sideways}\textit{SIP}\end{sideways}}& \multirow{4}{*}{\begin{sideways}\cite{fan2019rethinking}\end{sideways}} & $S_\alpha\uparrow$       & .850& .864& \textbf{.879}& .866& .870 & .855& .872 & .865\\
                                                                        && $F_{\beta}^{\textrm{max}}\uparrow$       & .851& .862& \textbf{.885}& .873& .873 & .857& .877 & .863\\
                                                                        && $E_{\phi}^{\textrm{max}}\uparrow$       & .903 & .910& \textbf{.923}& .916& .916 & .908& .920 & .913
\\                                                                     && $M\downarrow$       & .064 & .063& \textbf{.051}& .056& .055& .061& .056 & .061\\
\hline
\end{tabular}
\vspace{-0.4cm}
\end{table}

\textbf{RGB only $vs.$ depth only:} The comparison between columns ``\texttt{D}'' and ``\texttt{E}'' in Table \ref{tab_ablation} proves that using RGB data for saliency estimation is superior to using depth in most cases, indicating that the RGB view is generally more informative. However, using depth information achieves better results than RGB on SIP \cite{fan2019rethinking} and RGBD135 \cite{cheng2014depth}, as visualized in Fig. \ref{fig_ablationvisual}. This implies that the depth maps from the two datasets are of relatively good quality.

\textbf{Efficiency of JL component:}
Existing models usually use separate learning approaches to extract features from RGB and depth data, respectively. In contrast, our \ourmodel~adopts a joint learning strategy
to obtain the features from an RGB and depth map simultaneously.
We compare the two learning strategies and
%
find that using separate learning (two separate backbones) is likely to increase the training difficulties.
\figref{fig_learningcurve} shows typical learning curves for such a case. In the separate learning setting, where the initial learning rate is $lr=10^{-9}$, the network is easily trapped in a local optimum with high loss, while the joint learning setting (shared network) can converge nicely.
Further, for separate learning, if the learning rate is set to $lr=10^{-10}$, the learning process is rescued from local oscillation but converges slowly compared to our joint learning strategy. As shown in columns ``\texttt{B}'' and ``\texttt{F}'' in Table \ref{tab_ablation}, the resulting converged model after 40 epochs achieves worse performance than \ourmodel, namely 1.1\%/1.76\% overall drop on $S_{\alpha}$/$F_{\beta}^{\textrm{max}}$. We attribute the better performance of \ourmodel~to its joint learning from both RGB and depth data.

\section{Conclusion}\label{sec5}
\vspace{-5pt}
We present a novel framework for RGB-D based SOD, named \ourmodel, which is based on joint learning and densely-cooperative fusion.
Experimental results show the feasibility of learning a shared network for salient object localization in RGB and depth views, simultaneously, to achieve accurate prediction. Moreover, the densely-cooperative fusion strategy employed is effective for exploiting cross-modal complementarity. \ourmodel~shows superior performance against SOTAs on six benchmark datasets and is supported by comprehensive ablation studies. Our framework is quite flexible and general, and its inner modules could be replaced by their counterparts for further improvement.

\vspace{-2pt}
\small{\vspace{.1in}\noindent\textbf{Acknowledgments.}\quad
This work was supported by the NSFC, under No. 61703077, 61773270, 61971005, the
Fundamental Research Funds for the Central Universities No. YJ201755, and the Sichuan Science and Technology Major Projects (2018GZDZX0029).}

\clearpage

{\small
\bibliographystyle{ieee_fullname}
\bibliography{rgbd_egbib}
}

\end{document}